\pdfoutput=1

\documentclass[11pt]{article}
\usepackage{xcolor}
\usepackage[preprint]{acl}

\usepackage{times}
\usepackage{latexsym}

\usepackage[T1]{fontenc}

\usepackage[utf8]{inputenc}

\usepackage{microtype}

\usepackage{inconsolata}

\usepackage{graphicx}

\usepackage{subfigure}
\usepackage{booktabs}
\usepackage{multirow}

\usepackage{amsmath}
\usepackage{amssymb}
\usepackage{mathtools}
\usepackage{amsthm}

\usepackage{hyperref}

\usepackage[textsize=tiny]{todonotes}
\usepackage{xspace}

\usepackage{adjustbox}

\usepackage{tcolorbox}  
\usepackage{listings}   
\usepackage{caption}

\usepackage[frozencache,cachedir=minted-cache]{minted}

\usepackage{float}

\usepackage{colortbl}
\usepackage{xcolor}

\newcommand{\WarriorMath}{\textsc{WarriorMath}~}

\title{WarriorMath: Enhancing the Mathematical Ability of Large Language Models with a Defect-aware Framework}

\author{Yue Chen\thanks{These authors contributed equally to this work.}\thanks{This work was done when Yue and Minghua were interns at Microsoft.}$^{1,2}$, Minghua He\footnotemark[1]\footnotemark[2]$^{1,2}$, Fangkai Yang$^{2}$, Pu Zhao$^{2}$, Lu Wang$^{2}$, Yu Kang$^{2}$, \\ \textbf{Yifei Dong$^{3}$, Yuefeng Zhan$^{2}$, Hao Sun$^{2}$, Qingwei Lin$^{2}$, Saravan Rajmohan$^{2}$, Dongmei Zhang$^{2}$}  \\
  $^{1}$Peking University,
  $^{2}$Microsoft,
  $^{3}$KTH Royal Institute of Technology\\
}

\begin{document}
\maketitle


\begin{abstract}
Large Language Models (LLMs) excel in solving mathematical problems, yet their performance is often limited by the availability of high-quality, diverse training data. Existing methods focus on augmenting datasets through rephrasing or difficulty progression but overlook the specific failure modes of LLMs. This results in synthetic questions that the model can already solve, providing minimal performance gains.
To address this, we propose WarriorMath, a defect-aware framework for mathematical problem solving that integrates both targeted data synthesis and progressive training. In the synthesis stage, we employ multiple expert LLMs in a collaborative process to generate, critique, and refine problems. Questions that base LLMs fail to solve are identified and iteratively improved through expert-level feedback, producing high-quality, defect-aware training data. In the training stage, we introduce a progressive learning framework that iteratively fine-tunes the model using increasingly challenging data tailored to its weaknesses.
Experiments on six mathematical benchmarks show that WarriorMath outperforms strong baselines by 12.57\% on average, setting a new state-of-the-art. Our results demonstrate the effectiveness of a defect-aware, multi-expert framework for improving mathematical ability.
\end{abstract}

\begin{figure}[t]
	\centering  
	\subfigbottomskip=5pt 
	\subfigcapskip=5pt 
	\subfigure[Existing data synthesis strategy. Seed datasets are collected, and an external LLM is prompted to augment or label them. However, the resulting problems are often too simple, leading to limited improvements in model performance.
    ]{
		\includegraphics[width=1.0\linewidth]{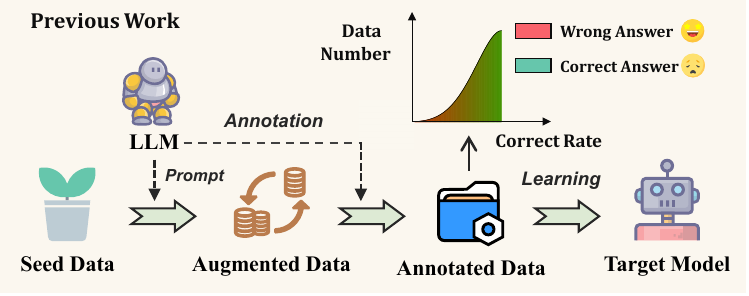}\label{teaser1}}
	\subfigure[Our defect-aware strategy. An Exam Committee of multiple expert LLMs generates, critiques, and refines problems, retaining only the failure data where the base LLMs struggled. This ensures synthesized problems challenge the base LLMs and enhance their capabilities.
    ]{
		\includegraphics[width=1.0\linewidth]{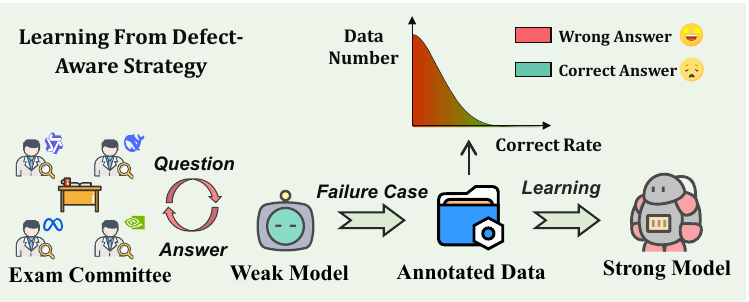}\label{teaser2}}
  \captionsetup{skip=2pt}
  \caption{
  Comparisons between our method and traditional data synthesis strategies.} 
  \label{teaser}
  \vspace{-0.5cm}
\end{figure}

\section{Introduction}
Large language models (LLMs) have demonstrated remarkable capabilities in solving mathematical and scientific problems~\citep{jaech2024openai, claude3.7sonnet, he2024llmelog, zeng2024matters, OpenAI2025o3mini, googlegemini252025, deepseekai2025deepseekr1}, positioning them as valuable mathematical assistants. Consequently, enhancing their mathematical ability has become a key research goal. 
~\citet{yang2024qwen25math} improve math skills through large-scale pre-training on math data, while ~\citet{muennighoff2025s1, wen2025lightr1, min2024imitate} focus on fine-tuning with high-quality instruction datasets. However, both approaches heavily rely on high-quality data~\citep{xu2023wizardlm, feng2025warriorcoder}, and key training data for strong models (e.g., OpenAI o1~\citep{OpenAI2025o3mini}) remain private, limiting reproducibility. Thus, collecting and annotating high-quality math problems at scale remains a major bottleneck.

Recent research has explored large-scale data synthesis using LLMs to enhance training datasets~\citep{tangmathscale,huang2024KPDDS,yue2024mammoth2scalinginstructionsweb,liu2024augmenting,zhou2024jiuzhang3, zhou2025trustragenhancingrobustnesstrustworthiness, li2024common7blanguagemodels,he2025execoder, luo2025wizardmath,li2024mugglemath, mei2025a1}. Some methods focus on mining instruction data from pretraining corpora ~\citep{yue2023mammothbuildingmathgeneralist,li2024selfalignmentinstructionbacktranslation}. Others generate data by rephrasing existing problems~\citep{yu2023metamath} or through difficulty progression~\citep{xu2023wizardlm, luo2025wizardmath}. 

However, these data synthesis approaches are not tailored to the base LLM, producing mostly solvable data that disregards inherent defects and offers little learning benefit (see Figure~\ref{teaser}). \emph{Inherent defects} refer to systematic failure (e.g., misinterpreting quantifiers, incorrect symbolic steps)~\citep{pan2025lemma, an2024learnfailure}. Unlike difficulty, defects reveal internal model limitations. Ignoring them leads to synthetic questions the model already solves, limiting improvement ~\citep{yu2025rethink, pan2025lemma, an2024learnfailure}. 
Existing research demonstrates that aligning training strategies with the model's evolving defects can significantly boost performance on challenging mathematical problems \citet{an2024learnfailure, wen2025lightr1, yu2025rethink, Wizardlm2}.
Therefore, a more interactive synthesis process that adapts to the model’s current performance is needed to effectively address these defects.

Based on this insight, we propose WarriorMath, a defect-aware framework for mathematical problem solving that integrates both data synthesis and progressive training, as shown in Figure~\ref{Fig:method}. WarriorMath decomposes data synthesis into two stages: \textbf{(1) Defect-aware problem construction.} We create an exam committee of state-of-the-art mathematical LLMs, each leveraging its expertise to design problems.  Judges then assess these problems' quality, while the base LLM provides feedback, identifying the knowledge it has yet to master. \textbf{(2) Answer generation and refinement.} Each model attempts to solve the problems with responses filtered and ranked through a combination of Elo rating and voting. This ensures that the solutions can effectively guide the base LLM's learning. Unlike prior approaches that expand existing datasets, WarriorMath generates novel, defect-specific training examples from scratch, enabling more efficient model improvement.

For training, we introduce a \textbf{progressive learning} framework that systematically addresses the model's defects through a two-stage process. 
WarriorMath first performs supervised fine-tuning (SFT) from answers generated by multiple expert models to absorb a broad foundation of mathematical knowledge. We then identify failure problems which the model still gets wrong and fine-tune the model to prefer stronger solutions on these samples. This process is repeated iteratively, allowing the model to correct its inherent defects without overriding previously mastered concepts, thereby enabling more effective learning from defects.

To evaluate the effectiveness of WarriorMath, we conduct evaluations  
on six prevalent mathematical benchmarks ~\citep{aime24data,aime25data, amc23data, lightman2023let, Minervadata, olympiadbenchdata}. Evaluation on these benchmarks indicates that WarriorMath achieves SOTA performance, surpassing existing same-sized open-source large models by an average of 12.57\%. Notably, the ablation experiments demonstrate that the proposed synthesis strategy can indeed generate proper high-quality data for the base LLM, as well as the effectiveness of the proposed training framework in learning from defects.

The key contributions of this work include:
\begin{itemize}
    \item We propose a defect-aware data synthesis pipline, which generate proper and high-quality data designed for base LLMs from scratch which emulates the educational philosophy of teaching according to aptitude. 
    \item We introduce a progressive learning framework that first learns broadly from experts and then improve ability through iterative alignment focusing on reinforcing knowledge where the model fails while bypassing mastered knowledge. 
    \item We demonstrate that \textbf{WarriorMath} achieves state-of-the-art performance among open-source LLMs, with strong data efficiency and generalization, validating the effectiveness of our approach.
\end{itemize}
\section{Related Work}
\subsection{Math LLMs}
Recent advances in LLMs’ mathematical capabilities have drawn growing attention from both academic and industrial communities. Early successes were fueled by the creation of large-scale pretraining corpora and curated fine-tuning datasets~\citep{paster2023openwebmathopendatasethighquality,wang2024mathpilebilliontokenscalepretrainingcorpus,shao2024deepseekmath,yue2023mammothbuildingmathgeneralist}, which significantly improved model accuracy on standard math benchmarks. This progress was further accelerated by specialized prompting strategies~\citep{cot,imani2023mathprompter}, tool augmentation~\citep{gao2023pal,schick2024toolformer}, and reinforcement learning techniques~\citep{deepseekai2025deepseekr1, zhao2024sego}.
While advanced prompting and test-time scaling methods~\citep{wu2024inference,muennighoff2025s1} continue to push performance limits, current LLMs still lag behind those of proprietary ones (e.g. GPT-4o, Claude, etc.), primarily because stronger models often keep their training data proprietary \citep{hui2024qwen25coder}. As a result, the lack of publicly available high-quality, diverse datasets remains a significant barrier to further development in this field.

\subsection{Data Synthesis}
Synthetic data has been employed to augment training datasets for various mathematical LLMs~\citep{luo2025wizardmath,Wizardlm2,li2024common7blanguagemodels}. Early approaches follow the Self-Instruct paradigm~\citep{wang2023self}, using few-shot prompting to generate synthetic instructions. To enhance diversity and difficulty, recent work~\citep{luo2025wizardmath, an2024learnfailure, liu2024augmenting} further explores iterative refinement and instruction evolution based on reasoning trajectories. While these methods improve the quality and diversity of synthetic data, they primarily focus on difficulty progression rather than addressing the model’s actual capability defects. That is, increasing difficulty does not necessarily target the failure cases of base LLMs. As a result, many synthesized questions remain solvable by the model, providing limited value for improving model ability. Moreover, many of these approaches depend heavily on proprietary LLMs (e.g., GPT-4, Claude)~\citep{muennighoff2025s1}, making large-scale data generation costly and less reproducible. Recent work such as \citet{feng2025warriorcoder} proposes a novel paradigm that distills data through multi-agent competitions among open LLMs, reducing reliance on external APIs. However, these methods still overlook the importance of targeting model-specific defects during synthesis.


\subsection{Learning From Defect}
An emerging line of work investigates how failure cases can be leveraged to guide LLMs toward improved performance. Reflexion~\citep{shinn2023reflexionlanguageagentsverbal} introduces a self-reflection mechanism in which the model analyzes past failures using either internal reflections or external feedback. Similarly, \citet{gou2024criticlargelanguagemodels} use external tools to provide real-time critiques, while \citet{chen2023teachinglargelanguagemodels} enable models to execute and debug code to enhance factual consistency. In contrast to these tool-based feedback approaches, WizardLM-2~\citep{Wizardlm2} proposes the AI-Align-AI (AAA) framework, where multiple LLMs collaboratively teach and critique each other. This setup involves simulated dialogues, quality evaluations, and constructive suggestions for improvement, allowing models to iteratively refine their outputs in a multi-turn process.
Despite these advances, most methods still treat failure feedback as a post-hoc augmentation rather than an integral part of the training process. Furthermore, few approaches systematically incorporate failure-driven supervision into dataset synthesis pipelines, leaving a gap between training data construction and the model’s evolving weaknesses.

\begin{figure*}[t!]
	\centering
	\includegraphics[width=1\linewidth]{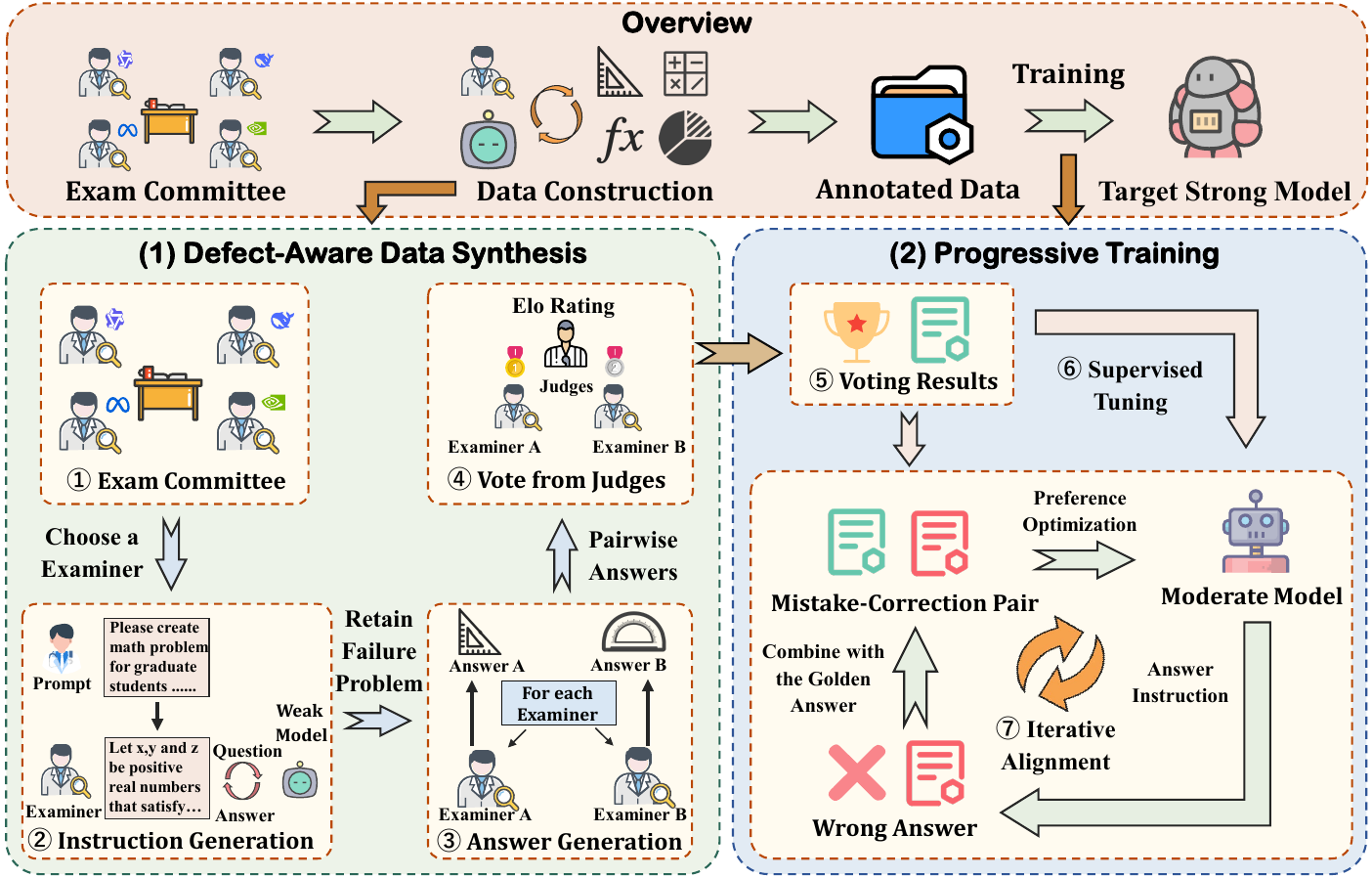}
    \caption{\textbf{The WarriorMath's pipeline.} 
    The WarriorMath pipeline includes two key components: \textbf{Defect-Aware data synthesis} and \textbf{Progressive Training}. 
    First, defect-aware instructions are synthesized. An exam committee composed of multiple expert models generates these instructions using carefully designed prompts. The base model then answers the generated questions, retaining only the failed defect data. Next, the synthesized instruction data is answered by examiners from the committee. Judges accurately select the best answers using Elo ratings to construct training instructions.
    Second, learning from the defect-related instructions. 
    Rapid training is performed through supervised fine-tuning to obtain a moderate model. 
    Then the moderate model re-answers the training instructions, and failed cases will be paired with golden answers and used for preference optimization. 
    The alignment process of answering-pairing-preference optimization proceeds in an iterative cycle to fully overcome model defects. 
    }
    \label{Fig:method}
    \vspace{-0.4cm}
\end{figure*}

\section{Method}
In this section, we elaborate on the details of our WarriorMath. As illustrated in Figure ~\ref{Fig:method}, the pipeline mainly contains two components: Defect-aware data synthesis and Progressive training. The details of data synthesis will be presented in \textsection~\ref{sec:synthesis} and the training method will be described in \textsection~\ref{sec:train}
\subsection{Defect-aware data synthesis}\label{sec:synthesis}
In the synthesis stage, we employ multiple expert LLMs in a collaborative process to generate, critique, and refine problems. Questions that base models fail to solve are identified and iteratively improved through expert-level feedback, producing high-quality, targeted training data.
\subsubsection{Problem generation}\label{sec:problem_gen}
\paragraph{Committee members Setting.} 
The capabilities of committee members determine the final performance of \textbf{WarriorMath}. Theoretically, the more diverse and high-quality training data are derived from a larger and stronger pool of members. For this study, we select five leading open-source math LLMs, DeepSeek-R1-Distill-Llama-70B~\citep{deepseekai2025deepseekr1}, Qwen2.5-Math-72B-Instruct~\citep{yang2024qwen25math}, QwQ-32B~\citep{qwq32b}, AceMath-72B-Instruct~\citep{liu2025acemath} and Phi-4-reasoning~\citep{abdin2025phi4reasoningtechnicalreport}. Notably, while \WarriorMath achieves state-of-the-art performance based solely on open-source Math LLMs, it can also learn from powerful proprietary LLMs. In each round of the data synthesis, only one math expert is selected as the examiner, while the remaining ones serve as judges.

\paragraph{Problem Synthesis from Scratch}
Considering each round of the data synthesis process between examiner LLM A and judge LLM B. The initial goal of this is to leverage LLM A’s capabilities to pose challenging questions to LLM B. To achieve this, we design a prompt that instructs LLM A to act as a world-class expert in generating difficult and diverse mathematical problems. Specifically, the prompt emphasizes four dimensions: (1) Quality, requiring problems to be clear, well-structured, and unambiguous; (2) Difficulty, demanding deep mathematical reasoning beyond pattern matching; (3) Diversity, covering a broad set of domains such as algebra, calculus, discrete mathematics, and geometry; and (4) Challenge, encouraging adversarial elements like subtle traps or misleading intermediate steps. This deliberate prompt design ensures LLM A activates its mathematical capabilities while producing maximally diagnostic questions to evaluate LLM B. Further Problem Synthesis details are provided in Appendix \ref{sec:synthesis-appendix}.

\paragraph{Deduplication and Defect-Aware assessment.}
To ensure the quality and informativeness of selected problems, we identify and filter out problems that are repetitive, ambiguous, or excessively difficult for the base LLM to learn from effectively. The Prompt for selection and discarded problems are provided in appendix ~\ref{sec:discard}. To further ensure that the retained problems are truly valuable for learning, we conduct a defect-aware assessment of the model’s performance. For each problem, we generate $N=16$ rollouts using the current model. After verifying the correctness of these outputs, we discard problems for which all sampled solutions are correct, retaining only those that the model answers incorrectly. Through interactive assessment, we confirm that these retained problems continue to provide meaningful learning signals.
Finally, to preserve both the diversity and representativeness of the instruction set, we apply the KCenterGreedy algorithm~\citep{sener2018active} to select a final subset $\bar{\bar{I}}$, using the \textit{all-roberta-large-v1} embedding model~\citep{liu2019robertarobustlyoptimizedbert} to compute semantic similarity between instructions.

\subsubsection{Answer Generation and refinement}
The reviewer is required to answer to the examiner’s question, while the reviewer A must also provide an answer to its own instruction (\textbf{Answer Generation}). Once both answers are collected, the judges (i.e., the remaining LLMs in the Committee ) will evaluate the correctness and helpfulness of the pairwise responses and vote for their preferred one. In order to assemble fair and accurate voting result, we introduce elo rating to overcome bias from LLM judges, inspired by \citep{bai2022train}. (more details about can be found in Appendix\ref{sec:vote_and_elo}).

\subsection{Progressive Training}\label{sec:train}
We propose a \textbf{Progressive Training} framework designed to iteratively enhance the model’s mathematical reasoning by systematically identifying and correcting its failure cases. This framework comprises two stages: supervised fine-tuning (SFT) and iterative alignment.

\paragraph{Stage 1: Supervised Fine-Tuning.}
Given a dataset, 
$\mathcal{D} = \{(x_i, Y_i, \{r_i^j\}_{j=1}^N)\}_{i=1}^N$, 
where $x_i$ is an instruction, 
$Y_i = \{y_i^j\}_{j=1}^N$ denotes the set of responses from different expert models, and $r_i^j$ is the score assigned to each response, we select the highest-scoring response as the gold standard:
\begin{equation}
y_i^{\text{gold}} = \arg\max_{y_i^j \in Y_i} r_i^j.
\label{eq:gold-selection}
\end{equation}
These high-quality instruction–response pairs $(x_i, y_i^{\text{gold}})$ are used to initialize the model $M_0$ via supervised fine-tuning using maximum likelihood estimation.

\paragraph{Stage 2: Iterative Alignment.}
To further refine the model, we adopt an iterative alignment strategy inspired by \citet{pang2024iterative}, where the model learns from its own mistakes by comparing incorrect outputs with expert references.

At iteration $t$, let $M_t$ be the current model. For each input $x_i$, we sample a set of $N_i$ responses:
\begin{equation}
G_i = \{(c_i^n, y_i^n)\}_{n=1}^{N_i} \sim M_t(x_i),
\label{eq:sampled-responses}
\end{equation}
where each response consists of a reasoning trace $c_i^n$ followed by a final answer $y_i^n$.

We then compute correctness labels for each response using a reward function $r_i^n = \mathcal{R}(y_i^n, \hat{y}_i)$, where $\hat{y}_i$ is the ground-truth answer. In practice, this reduces to simple matching:  

\begin{equation}
    r_i^n = 
\begin{cases}
1, & \text{if } y_i^n = \hat{y}_i \\
0, & \text{otherwise}
\end{cases}
\label{reward}
\end{equation}
This produces a labeled response set: 

\begin{equation}
G_i = \{(c_i^n, y_i^n, r_i^n)\}_{n=1}^{N_i}    
\end{equation}
From this set, we collect the incorrect responses:
\begin{equation}
G_i^{\text{neg}} = \{(c_i^n, y_i^n) \mid r_i^n = 0\}
\end{equation}

For each $(x_i, y_i^{\text{gold}})$ from Stage 1, we pair the gold reasoning-answer $(c_i^{\text{gold}}, y_i^{\text{gold}})$ with every incorrect model output $(c_i^l, y_i^l) \in G_i^{\text{neg}}$, forming preference training pairs:
\begin{equation}
\begin{aligned}
D_t^{\text{pairs}} = \big\{\, 
& \big((c_i^{\text{gold}}, y_i^{\text{gold}}), (c_i^l, y_i^l)\big) \;\big|\; \\
& x_i \in \mathcal{D},\; (c_i^l, y_i^l) \in G_i^{\text{neg}} 
\,\big\}.
\end{aligned}
\end{equation}
These preference pairs are used to optimize the model with a hybrid loss that combines the DPO (Direct Preference Optimization) loss and an auxiliary negative log-likelihood (NLL) loss:

\begin{equation}
\begin{aligned}
\mathcal{L}_{\text{total}} &= \mathcal{L}_{\text{DPO}} + \alpha \mathcal{L}_{\text{NLL}}, \\
\mathcal{L}_{\text{DPO}} &= - \log \sigma \Big( 
\beta \log \frac{M_\theta(c_i^\text{gold}, y_i^\text{gold} \mid x_i)}{M_t(c_i^\text{gold}, y_i^\text{gold} \mid x_i)} \\
&\quad - \beta \log \frac{M_\theta(c_i^l, y_i^l \mid x_i)}{M_t(c_i^l, y_i^l \mid x_i)} 
\Big), \\
\mathcal{L}_{\text{NLL}} &= - \frac{1}{|c_i^\text{gold}| + |y_i^\text{gold}|} 
\log M_\theta(c_i^\text{gold}, y_i^\text{gold} \mid x_i).
\end{aligned}
\label{eq:dopl-loss}
\end{equation}
where $\sigma$ is the sigmoid function, and $\alpha, \beta$ are balancing hyperparameters.
After optimization, we obtain the updated model $M_{t+1} = M_\theta$, which is then used in the next iteration to generate new responses and continue the correction process.

By repeatedly identifying and addressing its own errors, the model is gradually aligned with expert behavior. As a result, our approach progressively improves the model’s mathematical capability. In this way, \WarriorMath accumulates the expertise of diverse mathematical committee members, as encoded in their instructions and outputs throughout training.
\section{Experiment}
\subsection{Experimental Setup}
\paragraph{Backbones.} 
We implement WarriorMath with two initialization backbones: (1) WarriorMath-Qwen, initialized from Qwen2.5-Math-7B~\cite{yang2024qwen25math};
(2) WarriorMath-DS, initialized from DeepSeek-R1-Distill-Qwen-7B~\citep{deepseekai2025deepseekr1}. As for the competitors of expert battles, we choose strong open-source LLMs including DeepSeek-R1-Distill-Llama-70B~\citep{deepseekai2025deepseekr1}, Qwen2.5-Math-72B-Instruct~\citep{yang2024qwen25math}, QwQ-32B~\citep{qwq32b}, AceMath-72B-Instruct~\citep{liu2025acemath}, Phi-4-reasoning~\citep{abdin2025phi4reasoningtechnicalreport}.

\paragraph{Datasets.}
To evaluate the math capability of \textbf{WarriorMath}, we conduct evaluations  
on six prevalent mathematical benchmarks. Specifically, we use the following benchmarks:
(1) \textbf{AIME 2024, AIME 2025}~\citep{aime24data,aime25data} are benchmarks that include particularly challenging math problems from the American Invitational Mathematics Examination (AIME) of 2024 and 2025, designed to assess advanced problem-solving skills. 
(2) \textbf{AMC 2023:}~\citep{amc23data} The American Mathematics Competitions (AMC) are a series of examinations aimed at promoting the development of problem-solving skills. The 2023 AMC includes problems that test a range of mathematical concepts at the high school level.
(3) \textbf{MATH-500}~\citep{lightman2023let} is a dataset containing high school-level math problems. It serves to assess a model's ability to handle more advanced mathematical reasoning; and 
(4) \textbf{Minerva} ~\citep{Minervadata} is a benchmark dataset comprising a diverse collection of quantitative reasoning problems that cover topics such as arithmetic, algebra, geometry, calculus, physics, and chemistry, with difficulty levels ranging from grade school to college.
(5) \textbf{Olympiad Bench:} ~\cite{olympiadbenchdata} an Olympiad-level bilingual scientific benchmark, featuring 8,476 problems from Olympiad-level mathematics and physics competitions.
To ensure accurate evaluation, we follow the evaluation method proposed by~\citet{hochlehnert2025sober}. We use the Pass@1 metric as the primary evaluation criterion. For each result, we report the mean and standard deviation computed over multiple random seeds. All experiments are conducted using \texttt{lighteval}~\citep{lighteval} with the \texttt{vllm} backend~\citep{kwon2023efficient}.

\paragraph{Baseline.} 
We compare our method against reinforcement learning (RL) approaches trained on the Qwen2.5 Math Base models, including Oat-Zero~\citep{liu2025oatzero}, LIMR~\citep{li2025limr}, and
SimpleRL-Zoo~\citep{zeng2025simplerl}.
We also evaluate our approach against supervised fine-tuning (SFT) baselines, such as s1.1~\citep{muennighoff2025s1}, Eurus2 Prime~\citep{cui2025process}, Bespoke Stratos~\citep{bespoke-stratos-7b}, OpenR1~\citep{openr1} and OpenThinker~\citep{openthoughts}.
In addition, we consider recent state-of-the-art methods based on deepseek-r1-distill-qwen-7b as the backbone, such as LightR1~\citep{wen2025lightr1},and AReal-boba-RL-7b~\citep{AReal}.

\paragraph{Implentation Details} During the data synthesis, we adopt 9 different generation configs where temperature $t \in \{0.60,0.65,0.70\}$ and top-p $p \in \{0.85,0.90,0.95\}$. The detailed prompts can be found in Appendix~\ref{sec:prompt_ins}. As for the training stage, 
the global batch size is set to 512, and the number of total training steps is set to 448. We use a learning rate of \(1 \times 10^{-5}\) and a weight decay of \(3 \times 10^{-7}\). Additionally, a WarmupLR scheduler with a warmup ratio of 0.2 is used.
\begin{table*}[!ht]
\centering
\renewcommand{\arraystretch}{1.15}
\resizebox{1\textwidth}{!}{
\begin{tabular}{lccccccc}
\toprule
\textbf{Models} & \textbf{Base} & \textbf{AIME’24} & \textbf{AIME’25} & \textbf{AMC’23} & \textbf{MATH500} & \textbf{Minerva} & \textbf{Olympiad} \\
\midrule
\multicolumn{8}{c}{\textit{Qwen-Based Models}} \\
Qwen2.5-Math-7B-Base~\cite{yang2024qwen25math} & - & 20.7$\pm$3.8 & 8.7$\pm$3.9 & 56.2$\pm$5.7 & 64.3$\pm$0.5 & 17.3$\pm$1.9 & 29.0$\pm$0.5 \\
Qwen2.5-Math-7B-Instruct~\citep{yang2024qwen25math} & Qwen2.5-Math-7B & 15.7$\pm$3.9 & 10.7$\pm$3.8 & 67.0$\pm$3.9 & 82.9$\pm$0.1 & 35.0$\pm$0.6 & 41.3$\pm$0.9 \\
Qwen-2.5-Math-7B-SimpleRL-Zoo~\citep{zeng2025simplerl} & Qwen2.5-Math-7B & 22.7$\pm$5.2 & 10.7$\pm$3.4 & 62.2$\pm$3.6 & 76.9$\pm$1.8 & 30.1$\pm$2.8 & 39.3$\pm$0.6 \\
Qwen2.5-Math-7B-Oat-Zero~\citep{liu2025oatzero} & Qwen2.5-Math-7B & 28.0$\pm$3.1 & 8.8$\pm$2.5 & 66.2$\pm$3.6 & 79.4$\pm$0.3 & 34.4$\pm$1.4 & 43.8$\pm$1.1 \\
LIMR~\citep{li2025limr} & Qwen2.5-Math-7B & 30.7$\pm$3.2 & 7.8$\pm$3.3 & 62.2$\pm$3.4 & 76.5$\pm$0.4 & 34.9$\pm$1.3 & 39.3$\pm$0.9 \\
Qwen2.5-7B-Instruct~\citep{qwen25technicalreport} & Qwen2.5-7B & 12.3$\pm$3.2 & 7.3$\pm$3.4 & 52.8$\pm$4.8 & 77.1$\pm$1.2 & 34.9$\pm$1.0 & 38.7$\pm$1.0 \\
s1.1-7B~\citep{muennighoff2025s1} & Qwen2.5-7B & 19.0$\pm$3.2 & 21.0$\pm$5.5 & 59.5$\pm$3.7 & 80.8$\pm$0.6 & 37.5$\pm$1.1 & 48.2$\pm$1.4 \\
Eurus-2-7B-PRIME~\citep{cui2025process} & Qwen2.5-7B & 17.8$\pm$2.2 & 14.0$\pm$1.7 & 63.0$\pm$3.9 & 80.1$\pm$0.1 & 37.5$\pm$1.0 & 43.9$\pm$0.3 \\
Bespoke-Stratos-7B~\citep{bespoke-stratos-7b} & Qwen2.5-7B & 20.3$\pm$4.3 & 18.0$\pm$4.8 & 60.2$\pm$4.9 & 84.7$\pm$0.5 & 39.1$\pm$1.3 & 51.9$\pm$1.1 \\
\rowcolor{gray!20} WarriorMath-Qwen-7B & Qwen2.5-7B & \textbf{48.3$\pm$2.6} & \textbf{36.5$\pm$5.0} & \textbf{83.0$\pm$2.5} & \textbf{88.3$\pm$1.4} & \textbf{41.2$\pm$2.8} & \textbf{52.1$\pm$0.8} \\
\midrule
\multicolumn{8}{c}{\textit{DeepSeek-Based Models}} \\
DeepSeek-R1-Distill-Qwen-7B~\citep{deepseekai2025deepseekr1} & - & 52.3$\pm$6.3 & 39.0$\pm$5.9 & 91.5$\pm$2.7 & 94.1$\pm$0.3 & 40.1$\pm$0.4 & 67.3$\pm$0.1 \\
AReaL-boba-RL-7B~\citep{AReal} & DeepSeek-R1-Distill-Qwen-7B & 56.7$\pm$9.2 & 40.0$\pm$9.1 & 90.0$\pm$4.8 & 94.4$\pm$1.0 & 40.8$\pm$3.0 & 68.4$\pm$1.8 \\
\rowcolor{gray!20} WarriorMath-DS-7B & DeepSeek-R1-Distill-Qwen-7B & \textbf{60.0$\pm$9.1} & \textbf{50.7$\pm$9.1} & \textbf{93.2$\pm$4.9} & \textbf{95.0$\pm$1.0} & \textbf{43.20$\pm$1.9} & \textbf{69.6$\pm$1.8} \\
\bottomrule
\end{tabular}
}
\caption{\textbf{Evaluation results across six mathematical reasoning benchmarks.}We report Pass@1 accuracy (mean $\pm$ std) of all models across six math benchmarks under a standardized evaluation setup---results are averaged over ten seeds for AIME and AMC, and three seeds for the rest.}
\vspace{1mm}
\label{tab:overall}
\end{table*}

\begin{table*}[!ht]
\centering
\renewcommand{\arraystretch}{1.15}
\resizebox{0.70\textwidth}{!}{
\begin{tabular}{lcccc}
\toprule
\textbf{Models} & \textbf{Base} & \textbf{GSM8K} & \textbf{MATH-500} & \textbf{AIME2024} \\ 
\midrule
Qwen2.5-Math-7B-Instruct~\cite{yang2024qwen25math} & - & 95.2 & 83.6 & 13.3 \\ 
Openmathinstruct-7B~\cite{toshniwal2024openmathinstruct} & Qwen2.5-Math-7B & 92.0 & 79.6 & 10.0 \\ 
NuminaMath-7B~\cite{li2024numinamath} &Qwen2.5-Math-7B & 92.9 & 81.8 & 20.0 \\ 
Evol-Instruct-7B~\cite{luo2025wizardmath} &Qwen2.5-Math-7B & 88.5 & 77.4 & 16.7 \\ 
KPDDS-7B~\cite{huang2024KPDDS} &Qwen2.5-Math-7B & 89.9 & 76.0 & 10.0 \\ 
PROMPTCOT-Qwen-7B~\cite{zhao2025promptcot} &Qwen2.5-Math-7B & 93.3 & \textbf{84.0} & 26.7 \\ 
\rowcolor{gray!20} WarriorMath-Qwen-7b-SFT &Qwen2.5-7B & \textbf{95.7} & 83.8 & \textbf{36.7} \\
\bottomrule
\end{tabular}
}
\caption{Performance of different data synthesis strategies on three mathematical benchmarks.}
\label{tab:data_quality}
\end{table*}
\subsection{Performance and Comparison}
\paragraph{Main Result.}
The results on the math benchmarks are
summarized in Table~\ref{tab:overall}. WarriorMath achieves SOTA performance, with a pass@1 accuracy of 60\% in AIME’24
and 56.7\% in AIME'25, surpassing all other fine-tuned models. This highlights the efficacy of our approach in generating high-quality data and effective training process.
\begin{table*}[h]
    \centering
    \resizebox{1.0\linewidth}{!}{
    \begin{tabular}{ccc} \hline
         \textbf{Mathematics Domain} & \textbf{Percentage (\%)} & \textbf{Definition} \\
         \hline
         Applied Mathematics & 11.4 & Application of mathematical methods to solve problems in various fields like physics and engineering. \\
         Algebra & 30.3 & Study of mathematical symbols and rules for manipulating these symbols. \\
         Discrete Mathematics & 12.9 & Study of mathematical structures that are fundamentally discrete, such as graphs and integers. \\
         Geometry & 14.7 & Branch of mathematics concerned with the properties and relations of points, lines, surfaces, and solids. \\
         Number Theory & 13.1 & Study of integers and integer-valued functions. \\
         Precalculus & 1.2 & Mathematical preparation for calculus, covering topics like functions and trigonometry. \\
         Calculus & 1.8 & Study of continuous change, encompassing derivatives and integrals. \\
         Differential Equations & 0.5 & Equations involving derivatives that describe how quantities change. \\
         \hline
    \end{tabular}
    }
    \caption{The proportion of different tasks in the training data.}
    \label{tab:math_domains}
\end{table*}

\paragraph{Data Quality.}
In order to assess the effectiveness of the problem generation pipeline, we compare our method with the following problem generation baselines:
(1) \textbf{Evol-Instruct}: This method~\citep{luo2025wizardmath} aims to enhance the quality of instruction data by improving both its complexity and diversity, thus facilitating the generation of more varied and challenging problems; 
(2) \textbf{KPDDS}: A data synthesis framework~\citep{huang2024KPDDS} that generates question-answer pairs by leveraging key concepts and exemplar practices derived from authentic data sources; (3) \textbf{OpenMathInstruct}: This method~\citep{toshniwal2024openmathinstruct} utilizes few-shot learning to prompt an LLM to create new math problems based on existing examples, without explicit instructions for adjusting difficulty or introducing new constraints; (4) \textbf{NuminaMath}: This approach~\citep{li2024numinamath} uses an LLM to generate novel math questions starting from a reference problem; (5) \textbf{PROMPTCOT}: This method~\citep{zhao2025promptcot} synthesizes complex problems based on mathematical concepts and the rationale behind problem construction, emulating the thought processes of experienced problem designers. For fair comparisions we follow the evaluation scripts provided in ~\citep{zhao2025promptcot}. 

The results of data quality assessment, presented in Tables~\ref{tab:data_quality},
Our method achieves state-of-the-art performance across multiple benchmarks, outperforming the
baselines, which highlights the efficacy of our defect-aware 
approach in generating high-quality problems

\paragraph{Iterative Defect Alignment.}
As a seed model $M_0$ we use the WarriorMath-Qwen-7b-SFT, which is fine-tuned with instruction data generated by Defect-Aware Committee Assessment. In each iteration, we generate $N=32$ solutions per problem using sampling with temperature 0.7 and top-p 0.8, and verify the answer to select wrong solution $(c_i, y_i)$ in the loser set $G_i^l$. Then we generate $K=10$ pairs per problem for training with our loss in \autoref{eq:dopl-loss}. In total, we perform three iterations, producing models $M_1$, $M_2$, $M_3$. The coefficient $\alpha$ is tuned in \{0.25, 0.5, 1\} when training $M_1$, and we end up using 1 for all experiments in the paper. The coefficient $\beta$ in the DPO loss is tuned in \{0.05, 0.1, 0.5\}, and we end up using 0.1 in this experiment. 
\begin{table}[t!]
    \centering
    \resizebox{0.8\linewidth}{!}{
    \begin{tabular}{lc}
        \toprule
        \textbf{Model} & \textbf{AIME'24} \\
        \midrule
        WarriorMath-Qwen-7b-SFT & 36.7$\pm$4.5 \\
        \quad \emph{Iteration 1} & 42.5$\pm$5.5 \\
        \quad \emph{Iteration 2} & 44.7$\pm$7.2 \\
        \quad \emph{Iteration 3} & \textbf{48.3$\pm$2.6} \\
        \bottomrule
    \end{tabular}
    }
    \vspace{-1mm}
    \caption{Results with Iterative Alignment.}
    \label{tab:iter_num}
\end{table}

{Overall results are given in \autoref{tab:iter_num}. We find that WarriorMath outperforms supervised fine-tuning (SFT) on the gold (dataset-provided) data, and steady growth over the iteration rounds.
\subsection{Ablation Study.}
\subsubsection{Number of Committee Members}
Table~\ref{tab:expert_num} presents the results observed when the target model learns from varying numbers of committe members. The target model shows a significant improvement when learning from just one mathematical LLM, indicating that even a single committe member enables it to acquire a specific set of knowledge. However, as the number of members increases, \textbf{WarriorMath} benefits from learning across all mathematical LLMs. As a result, the model trained with 5 math LLMs outperforms others across all 6 benchmarks, demonstrating the advantages of integrating knowledge from multiple specialized experts.

\begin{table}[t!]
    \centering
    \renewcommand{\arraystretch}{1.15}
    \resizebox{0.8\linewidth}{!}{
    \begin{tabular}{cccc} \hline
         {\textbf{\#Num}}& {\textbf{AIME'24}}&  {\textbf{AIME'25}}& {\textbf{AMC'23}} \\
         \hline
         {1}& {19.7$\pm$2.9}&  {15.7$\pm$2.7}&  {59.5$\pm$4.5} \\
         {2}& {29.4$\pm$3.2}&  {23.5$\pm$4.2}&  {69.3$\pm$3.6} \\
         {5}& {48.3$\pm$2.6}&  {36.5$\pm$5.0}&   {83.0$\pm$2.5} \\
         \hline
    \end{tabular}
    }
    \caption{The results observed when learning from varying numbers of committee members.}
    \vspace{-3mm}
    \label{tab:expert_num}
\end{table}
\subsubsection{Data Analysis}
\paragraph{Data Dependence}
As mentioned in \ref{sec:problem_gen}, our data sources are from multi-expert LLMs, rather than common resources frequently adopted by other datasets. To evaluate the data
novelty and dependence of ours data, we performed the following analysis for all the datasets. Figure~\ref{Fig:rouge} illustrates the overlap between the instructions mined from expert LLMs and those from two widely used math training datasets: (1) DeepScaleR~\citep{deepscaler2025} and (2) Omni-MATH~\citep{gao2024omnimathuniversalolympiadlevel}, measured using the ROUGE score. The majority of the mined instructions have a ROUGE score below 0.3, indicating that they are largely distinct from those in existing datasets. Notably, none of the mined instructions exceed a ROUGE score of 0.6, which further underscores that these instructions originate from the internal distribution of expert LLMs rather than being simple replications or extensions of the training data. As a result, these instructions demonstrate a higher degree of independence and are particularly valuable for training, as they provide novel examples that can enhance the capabilities of the target model.

\paragraph{Data Diversity}
A core attribute of our training data is its broad topical diversity across the mathematical domains. As shown in Table~\ref{tab:math_domains}, the classification results demonstrate that the data synthesis by WarriorMath encompasses tasks from numerous core mathematical domains. Its coverage extends from elementary subjects such as Applied Mathematics and Basic Geometry to advanced fields like Number Theory and Differential Equations. This extensive and profound topical base guarantees that models trained on WarriorMath are immersed in a rich array of mathematical concepts and problem-solving frameworks, thus facilitating the cultivation of more robust and widely applicable  mathematical capabilities.

\section{Conclusion}
In this work, we present WarriorMath, a defect-aware framework that enhances mathematical ability in LLMs through defect-aware data synthesis and progressive training. Our method constructs high-quality, defect-aware training data by leveraging a committee of expert LLMs to generate, critique, and refine problems specifically designed to expose the base LLM’s inherent defects. Through a two-stage progressive training process, WarriorMath incrementally aligns the model to stronger mathematical ability. Extensive experiments on six mathematical benchmarks demonstrate that WarriorMath achieves state-of-the-art performance among open-source models, highlighting the importance of learning from model-specific defects. 

\section*{Limitation}
In this paper, we introduce a novel training paradigm where the target model learns through Defect-Aware data synthesis, effectively addressing the limitations of existing data flywheels. This approach allows for the low-cost generation of high-quality and diverse data from scratch. However, as the number of expert models increases, the evaluation process becomes increasingly time-consuming. Designing more efficient and scalable multi-agent collaboration mechanisms remains an important direction for future research.

\bibliography{references}

\newpage
\appendix
\onecolumn
\section{Instruction data synthesis}\label{sec:synthesis-appendix}
\label{sec:ex_con}

\subsection{Prompts for Instruction Mining}\label{sec:prompt_ins}
\begin{figure*}[htbp]
\begin{tcolorbox}[title=\textbf{Instruction Mining Prompt}]

\textbf{Prompt:} 

Please act as a world-class expert in designing extremely challenging and diverse math problems. Your goal is to create problems that thoroughly test a model's reasoning abilities by inducing a variety of potential failure modes (e.g., reasoning, understanding, calculation, or strategy errors).

For each problem you design, please ensure the following:

\begin{enumerate}
    \item \textbf{Quality:} Questions must be well-formatted, clearly structured, and unambiguous.
    \item \textbf{Difficulty:} Problems should require deep mathematical reasoning and not be solvable via simple pattern recognition or surface-level heuristics.
    \item \textbf{Diversity:} Problems must span a wide range of mathematical domains, such as algebra, calculus, discrete math, geometry, and others.
    \item \textbf{Challenge:} Each problem should be adversarially constructed to trigger potential weaknesses in advanced models, such as subtle traps or misleading intermediate steps.
\end{enumerate}
Always provide the final answer enclosed within \texttt{\textbackslash boxed\{\}} for clarity.
\end{tcolorbox}
\caption{The prompt for generating challenging and diverse math problems.}
\label{fig:math_generation_prompt}
\end{figure*}

\newpage
\subsection{Case study of mining problems}
\begin{figure*}[htbp]
\captionsetup{font=normalsize, skip=15pt}
\begin{tcolorbox}[
    title={\normalsize \textbf{Examples of mining problems}}
]
\linespread{1.1}\selectfont 

\textbf{Case \#1}: Let \( S \) be the set of all convex quadrilaterals inscribed in the circle \( x^2 + y^2 = 25 \) with vertices at points having integer coordinates. Determine the maximum possible area of such a quadrilateral \( Q \), given that the sum of the \( x \)-coordinates of its vertices equals the sum of the \( y \)-coordinates. Express your answer as a reduced fraction \( \frac{m}{n} \), where \( m \) and \( n \) are coprime positive integers, and find \( m + n \).

\hrulefill

\textbf{Case \#2}: Consider triangle \( ABC \) with \( AB = 13 \), \( BC = 14 \), and \( AC = 15 \). Let \( O \) be the circumcenter and \( H \) the orthocenter. Let the circumradius be \( R \). A circle centered at \( O \) with radius \( R/2 \) intersects the nine-point circle at points \( P \) and \( Q \). Find the length of \( PQ \).

\hrulefill

\textbf{Case \#3}: Consider the function \( f(x, y) = x^3 + y^3 - 3xy \). Find the maximum value of \( f(x, y) \) on the closed disk \( x^2 + y^2 \leq 1 \).

\hrulefill

\textbf{Case \#4}: Consider a sequence of \( n \) independent coin flips, where each flip has a probability \( p \) of landing heads and a probability \( q = 1 - p \) of landing tails. Let \( X \) be the random variable representing the number of heads in the sequence. Find the probability that \( X \) is even.

\hrulefill

\textbf{Case \#5}: Let \( f(x) = x^3 - 3x + 1 \). The polynomial \( f(x) \) has three real roots, denoted by \( \alpha, \beta, \gamma \). Define the sequence \( \{a_n\} \) by \( a_1 = \alpha + \beta + \gamma \), \( a_2 = \alpha^2 + \beta^2 + \gamma^2 \), and for \( n \geq 3 \), \( a_n = \alpha^n + \beta^n + \gamma^n \). Find the value of \( a_{2023} \) modulo 3.

\hrulefill

\textbf{Case \#6}: Consider the set \( S = \{1, 2, 3, \ldots, 100\} \). A subset \( A \) of \( S \) is called "sum-free" if there do not exist distinct elements \( a, b, c \in A \) such that \( a + b = c \). Determine the maximum possible number of elements in a sum-free subset of \( S \).

\hrulefill

\textbf{Case \#7}: A fair six-sided die is rolled repeatedly until a 6 appears. Let \( X \) be the number of rolls required. Define the function \( f(n) \) as the probability that \( X \) is a multiple of \( n \). Find the value of \( f(3) \).

\hrulefill

\textbf{Case \#8}: Determine the value of \( a_{100} \) for the sequence defined by \( a_1 = 1 \) and \( a_{n+1} = a_n + \gcd(a_n, n) \) for \( n \geq 1 \).
\end{tcolorbox}
\caption{Examples of mining problems.}
\label{fig:mining_problems}
\end{figure*}

\newpage
\subsection{Prompt for Problem quality evalution and Discarded Problems examples}\label{sec:discard} 
\begin{figure*}[h]
\linespread{0.9}\selectfont  
\begin{tcolorbox}[title=\textbf{Math Instruction Quality Evaluation Prompt}]
You are a senior university-level mathematics instructor with extensive expertise in advanced topics such as Algebra, Precalculus, Number Theory, Geometry, and Combinatorics. Your task is to \textbf{evaluate the quality of mathematical problem statements} based on their clarity, formatting, conceptual soundness, computational complexity, and contextual relevance. Each problem should be scored on a scale of \textbf{1 to 10}, and your output must follow a structured JSON format.
\textbf{Evaluation Criteria:}
\begin{itemize}
    \item \textbf{Clarity and Completeness:} Is the problem clearly stated without ambiguity? Are all necessary variables, conditions, and constraints defined? Is the mathematical notation properly used and well-structured?
    \item \textbf{Conceptual Soundness and Difficulty:} Does the problem involve meaningful, non-trivial mathematical reasoning or advanced concepts? Does it promote critical thinking and apply appropriate mathematical principles?
    \item \textbf{Computational Complexity:} Does the solution process require more than basic arithmetic or trivial computation? Are there non-obvious calculations, transformations, or logical deductions involved?
    \item \textbf{Contextual Relevance and Verifiability:} Is the problem well-grounded in a practical, educational, or theoretical context? Is the problem solvable or verifiable using existing tools and methods? Avoid problems that are vague, proof-based without criteria, or ill-posed.
\end{itemize}
\textbf{Scoring Scale:}
\begin{itemize}
    \item \textbf{Excellent (9–10):} The instruction is verifiable, properly formatted, and conceptually sound.
    \item \textbf{Good (6–8):} Minor issues in clarity or formatting. Still verifiable and mathematically valid.
    \item \textbf{Average (3–5):} Noticeable flaws in clarity, completeness, or relevance.
    \item \textbf{Poor (1–2):} Ambiguous, improperly defined, unverifiable, or conceptually flawed.
\end{itemize}
\textbf{Your Output Format:}
Your output must be a \textbf{JSON list}, where each element is a dictionary with the following keys:
\begin{itemize}
    \item \texttt{instruction}: The original math problem.
    \item \texttt{score}: An integer from 1 to 10 representing your evaluation.
    \item \texttt{reason}: A detailed explanation justifying the score based on the criteria above.
\end{itemize}
\end{tcolorbox}
\caption{Prompt used to evaluate the quality of mathematical instructions, including scoring criteria and output format. Only instructions rated 6 or higher are considered suitable for use in further steps.}
\label{fig:instruction_eval_prompt}
\end{figure*}

\newpage
\begin{figure*}[h]
\begin{tcolorbox}[title=\textbf{Examples of discarded problems}]
We provide examples of low-quality problems that were filtered out during problem selection, categorized according to our criteria (Verifiability, Proper Formatting, Clarity):

\begin{itemize}
    \item \textbf{(1) Unverifiable Problem:}
    
    \texttt{Write a python code that run a math function like "log(base , number)".}
    
    \textit{Reason: This problem is not a mathematical question with a concrete answer, and cannot be automatically evaluated.}
    
    \medskip
    
    \item \textbf{(2) Poor Formatting:}
    
    \texttt{With what polynomial function equation do you want to calculate the vertex of the graph? \\
    f(x)=2x<sup>2</sup>-x<sup>3</sup>+5x<sup>4</sup>}
    
    \textit{Reason: The input mixes natural language with improperly rendered HTML, leading to parsing and readability issues.}
    
    \medskip
    
    \item \textbf{(3) Incomplete Problem:}
    
    \texttt{Chef's portion took 15 seconds, and the assistant's portion took 45 seconds.}
    
    \textit{Reason: The question is incomplete and lacks a clear task or objective for the model to solve.}
\end{itemize}

\end{tcolorbox}
\caption{Examples of discarded problems during the filtering process.}
\label{fig:discarded_problems}
\end{figure*}

\newpage
\subsection{Vote and elo rating}\label{sec:vote_and_elo}
The reviewer B is required to respond to the Examiner’s question, while the Examiner A must also provide an answer to its own instruction. Then we can calculate the \textit{local score} for each response:
\begin{equation}
\begin{aligned}
x^i_{A>B} = \frac{t_{A}}{t_{A}+t_{B}} \quad x^i_{B>A} = \frac{t_{B}}{t_{A}+t_{B}}
\end{aligned}
\label{vote}
\end{equation}
where $x^i_{A>B}$ and $x^i_{B>A}$ are the local scores for A's and B's responses to the instruction $i$. $x^i_{A>B}$ represents the percentage of votes that candidate A receives, while $x^i_{B>A}$ similarly represents the percentage of votes that candidate B receives. $t_{A}$ and $t_{B}$ are the number of votes which A and B win.

However, relying solely on the \textit{local score} to select the winner can be problematic. In some cases, a weaker model may receive more votes than a stronger one, even though its responses are not significantly better. This can occur because the \textit{local score} may not fully capture the quality of the model’s performance, especially in situations where the voting is influenced by factors, such as randomness or bias from LLM judges.

To address this limitation, we propose considering both local contingency and global consistency in the decision-making process. Instead of directly basing our analysis on the immediate voting outcomes, we introduce the concept of the \textit{global score} — specifically, the Elo rating
, which provides a more comprehensive reflection of a model's relative performance over time and across various evaluations. The Elo rating system, originally developed to calculate the relative skill levels of players in two-player games (such as chess), has been successfully adapted to assess the performance of competitors in a range of competitive scenarios, including esports and other skill-based games.

By incorporating the Elo rating, we account for both local performance in individual contests and global performance across multiple rounds, providing a more robust and accurate measure of a model's overall ability. This helps to mitigate the risk of weak models winning based on isolated, potentially unrepresentative votes:
\begin{equation}
\begin{aligned}
&X^{Elo}_{A>B} = \frac{1}{1+{10}^{(R_B-R_A)/400}} \\
&X^{Elo}_{B>A} = \frac{1}{1+{10}^{(R_A-R_B)/400}} \\
\end{aligned}
\label{elo}
\end{equation}
where $X^{Elo}_{A>B}$ and $X^{Elo}_{B>A}$ indicate the expected probabilities of A defeating B and B defeating A, respectively. $R_A$ and $R_B$ are the Elo rating of A and B, which are updated dynamically and iteratively. Given the battle result of A and B on an instruction $i$, we update them by:
\begin{equation}
\begin{aligned}
R_A \gets R_A + K \times (s^i_{A>B}-X^{Elo}_{A>B}) \\
R_B \gets R_B + K \times (s^i_{B>A}-X^{Elo}_{B>A})
\end{aligned}
\label{update}
\end{equation}
where $s^i_{A>B}$ and $s^i_{B>A}$ are the actual score of the battle result of player A and B (1 for a win, 0.5 for a draw, and 0 for a loss). The factor $K$ controls the sensitivity of rating changes.

Based on Equation~\ref{vote} and Equation~\ref{elo}, we can obtain the final score of A's response for instruction $i$:
\begin{equation}
\begin{aligned}
e^{i}_{A} = \sum_{B \in Com \setminus A} \alpha X^{Elo}_{A>B} + (1-\alpha) x^i_{A>B}
\end{aligned}
\label{final}
\end{equation}

where $Com$ is the set of all the competitors and `$\setminus$' is the subtraction operation. $\alpha$ is the coefficient to balance the local contingency and global consistency.

\begin{figure*}[htbp]
\begin{tcolorbox}[title=\textbf{Pair-wise Answer Quality Evaluation Prompt}]

\textbf{Prompt:} 

Please act as an impartial judge and evaluate the quality of the response provided by an AI assistant to the user prompt displayed below.

You will be given a user prompt, a reference answer, and the assistant's answer. Your job is to compare the assistant's answer with the reference one and assign a score.

For each user prompt, carry out the following steps:

\begin{enumerate}
    \item Consider if the assistant's answer is helpful, relevant, and concise. 
    \begin{itemize}
        \item \textbf{Helpful} means the answer correctly responds to the prompt or follows the instructions. 
        \item \textbf{Relevant} means all parts of the response closely connect or are appropriate to what is being asked.
        \item \textbf{Concise} means the response is clear and not verbose or excessive.
    \end{itemize}
    
    \item Then consider the creativity and novelty of the assistant's answer when needed.
    
    \item Identify any missing important information in the assistant's answer that would be beneficial to include when responding to the user prompt.
    
    \item After providing your explanation, you must rate the assistant's answer on a scale of 1 to 10, where a higher score reflects higher quality.
\end{enumerate}

\textbf{Guidelines for Scoring:}
\begin{itemize}
    \item \textbf{Assistant's Answer >> Reference Answer (7--10):} The assistant's answer is significantly or slightly better than the reference answer.
    \item \textbf{Assistant's Answer == Reference Answer (5--6):} The quality of assistant's answer is relatively the same as that of the reference answer.
    \item \textbf{Assistant's Answer << Reference Answer (1--4):} The assistant's answer is significantly or slightly worse than the reference answer.
\end{itemize}

\bigskip

\textbf{User Prompt:} \\
\texttt{\{instruction\}}

\medskip

\textbf{Reference Answer:} \\
\texttt{\{reference\}}

\medskip

\textbf{Assistant's Answer:} \\
\texttt{\{response\}}

\bigskip

Use double square brackets to format your scores, like so: \texttt{[[7]]}.

\end{tcolorbox}
\caption{The prompt for the evaluation of pairwise comparison.}
\label{fig:voting}
\end{figure*}

\newpage
\section{Data Analysis}
\subsection{Data Dependence}
\begin{figure}[H]
	\centering	\includegraphics[width=1.0\linewidth]{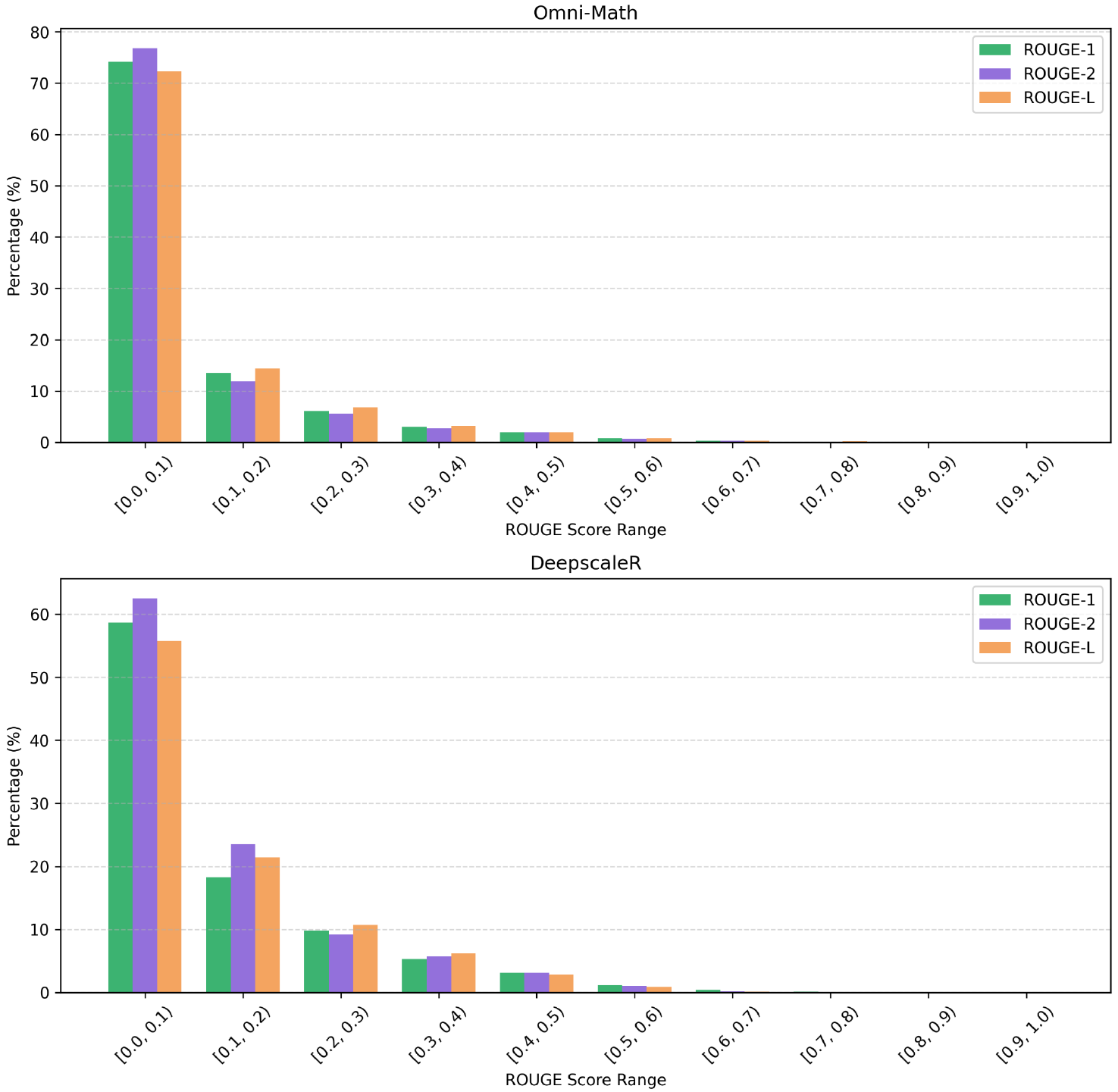}
    \caption{The overlapping rate between the mined instructions and existing training datasets.}
    \label{Fig:rouge}
\end{figure}

\end{document}